\pdfoutput=1
\documentclass[11pt]{article}

\usepackage[preprint]{coling}

\usepackage{times}
\usepackage{latexsym}

\usepackage[T1]{fontenc}
\usepackage[utf8]{inputenc}

\usepackage{microtype}

\usepackage{inconsolata}

\usepackage{graphicx}

  \usepackage{tabularray}
  \usepackage{amsmath}
  \usepackage{graphicx}
  \usepackage{subfig}
  \usepackage{hyperref}

\title{CoPrUS\@: Consistency Preserving Utterance Synthesis towards more realistic benchmark dialogues}

	\author{Sebastian Steindl \\
		Ostbayerische Technische \\ Hochschule Amberg-Weiden \\ Germany \\ s.steindl@oth-aw.de \\\And
		Ulrich Schäfer \\
		Ostbayerische Technische \\ Hochschule Amberg-Weiden \\ Germany \\ u.schaefer@oth-aw.de \\\And
		Bernd Ludwig \\
		University Regensburg \\ Germany \\
		bernd.ludwig@ur.de
	}

\begin{document}
\maketitle
\begin{abstract}
    Large-scale Wizard-Of-Oz dialogue datasets have enabled the training of deep learning-based dialogue systems.
    While they are successful as benchmark datasets, they lack certain types of utterances, which would make them more realistic.
    In this work, we investigate the creation of synthetic communication errors in an automatic pipeline. 
    Based on linguistic theory, we propose and follow a simple error taxonomy.
    We focus on three types of miscommunications that could happen in real-world dialogues but are underrepresented in the benchmark dataset: misunderstandings, non-understandings and vaguely related questions.
    Our two-step approach uses a state-of-the-art Large Language Model (LLM) to first create the error and secondly the repairing utterance.
    We perform Language Model-based evaluation to ensure the quality of the generated utterances.
    We apply the method to the MultiWOZ dataset and evaluate it both qualitatively and empirically as well as with human judges.
    Our results indicate that current LLMs can aid in adding post-hoc miscommunications to benchmark datasets as a form of data augmentation.
    We publish the resulting dataset\footnote{\url{https://github.com/sebastian-steindl/CoPrUS_data}}, in which nearly 1900 dialogues have been modified, as CoPrUS-MultiWOZ to facilitate future work on dialogue systems.
\end{abstract}

\section{Introduction}
  \begin{figure} [t]
    \includegraphics[width=\linewidth]{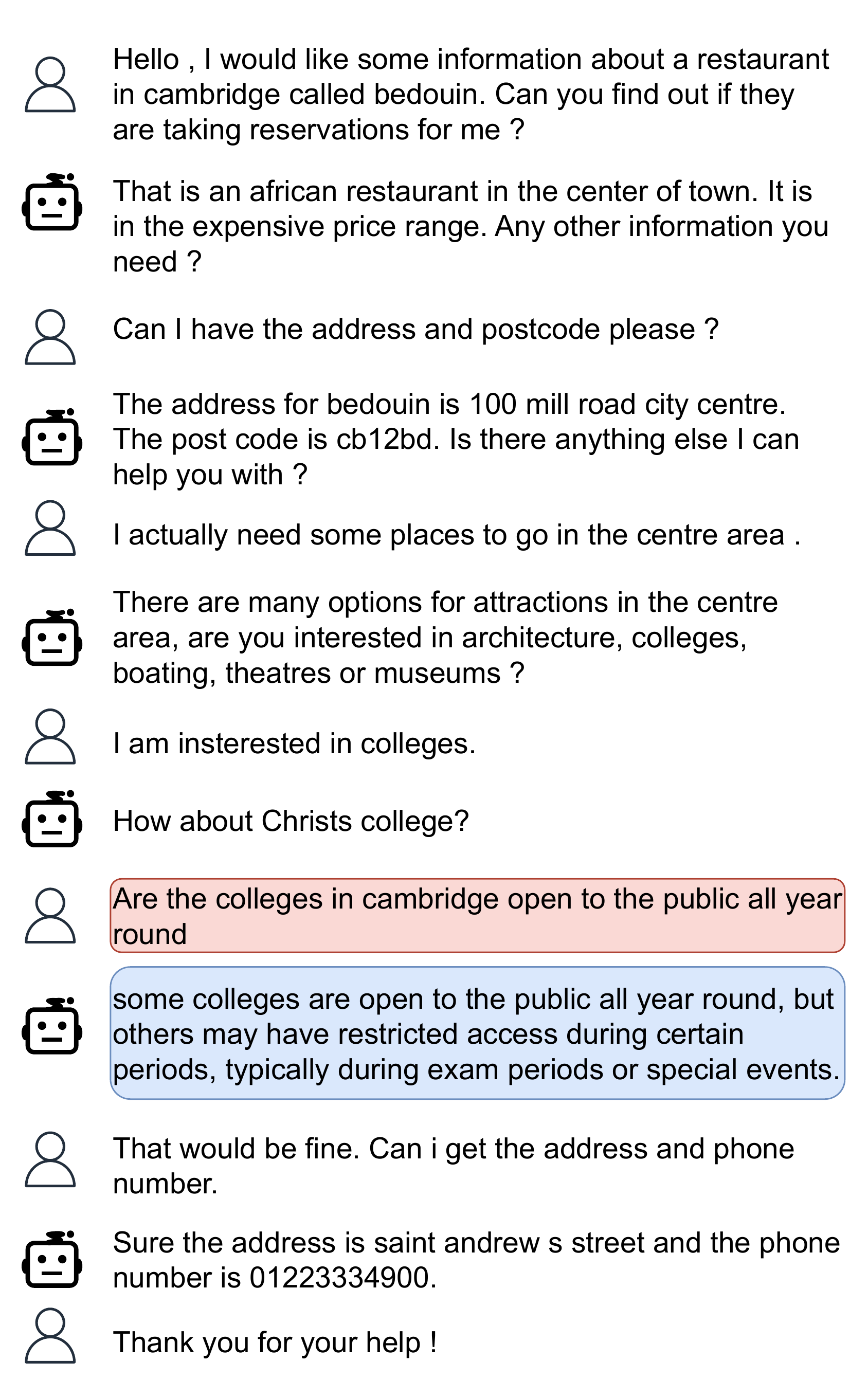}
    \caption{Example MultiWOZ dialogue after application of CoPrUS\@. Highlighted red is the miscommunication utterance, in this case, a vaguely related question, and in blue the repairing attempt by the system.}
    \label{fig:main}
  \end{figure}
  Task-Oriented Dialogue (TOD) systems are a special case of chatbots that help the user achieve certain tasks, such as booking a hotel, through conversation.
  This requires the TOD system to understand the user's utterance, decide on a policy, and utter a response to the user \cite{heGalaxyGenerativePretrained2022}.

  With the great advances in model architectures, these systems nowadays utilize deep learning and transformer models instead of being rule-based \cite{su-etal-2022-multi, bang-etal-2023-task, zhaoDescriptionDrivenTaskOrientedDialog2022}.
  This, however, requires large amounts of labeled training data. 
  The gold standard for collecting these dialogues is the Wizard-Of-Oz (WOZ) technique \cite{kelleyIterativeDesignMethodology1984}, in which two humans simulate a human-machine interaction. One widespread benchmark dataset that was created with Wizard-Of-Oz is MultiWOZ \cite{budzianowski-etal-2018-multiwoz}.

  Even though the dialogues were created by two humans, we find them \textit{sterile} and unrealistic. 
  More precisely, they are very close to what would be considered the idealistic \textit{happy path}, where everything works as expected without sidetracking. 
  Previous work has also identified multiple caveats within the MultiWOZ dataset, which we will discuss in section~\ref{sec:UnrealisticBenchmark}.
  Nonetheless, to make dialogue systems generalize well into real-world usage, it is of high priority that the training data is realistic, since the system will only be able to adequately answer utterances that are close to the training distribution.
 
  The current state-of-the-art Large Language Models (LLMs) are known for their strong capabilities in Natural Language Generation (NLG), instruction-following and in-context learning \cite{ouyangTrainingLanguageModels2022,sanh2022multitask}. 
  In this study, we thus investigate whether LLMs can aid in making task-oriented dialogue datasets more realistic by adding synthetic miscommunications. We see this as a form of data augmentation that is not focused on the quantity of the data, but on its quality in the sense of one of its properties, i.e., lack of miscommunications.

  We specifically focus on three types of miscommunication turns and their repairing, that are lacking in the benchmark: non-understanding, misunderstanding, and vaguely related questions. These should not change the goal or consistency of the dialogue, since their only motivation is clarification.
  The inclusion of these types of turns can make the dialogues more realistic and improve the ease of use for the dialogue system.
  This could allow for a higher level of technology acceptance~\cite{mitznerOlderAdultsTalk2010} and reduce frustration in all user age groups~\cite{vandergootExploringAgeDifferences2020}.
  
  We thus propose consistency-preserving utterance synthesis (CoPrUS) to make benchmark dialogues more realistic post-hoc.
  We evaluate our method empirically and with a qualitative analysis.
  Moreover, we publish the dataset that results from applying CoPrUS to the MultiWOZ 2.1~\cite{eric-etal-2020-multiwoz} dataset in the unified format of \citet{convlab3}. 

  \section{Background and Related Work}
  In the following, we will first give an overview of problems with the used TOD benchmark dataset and TOD systems. 
  Then, we will go into the differences between human-human and human-machine conversations and finally talk about taxonomies for conversational errors. 
  \label{sec:background}
 
  \subsection{Task-Oriented Dialogue Systems}
  Initially, TOD systems addressed each subtask (Natural Language Understanding, Policy Planning, NLG) with a specific (rule-based) model \cite{youngPOMDPBasedStatisticalSpoken2013}.
  Yet the publication of larger datasets, such as MultiWOZ~\cite{budzianowski-etal-2018-multiwoz}, in combination with general progress in deep learning, led to the usage of (end-to-end) deep learning approaches, e.g., as proposed by \citet{lin-etal-2020-mintl, peng-etal-2021-soloist, heGalaxyGenerativePretrained2022,sunMarsModelingContext2023, bang-etal-2023-task}.
  Lately, LLMs have also been investigated as zero-shot TOD systems \cite{ulmer-etal-2024-bootstrapping, li-etal-2024-large-language-models, chung-etal-2023-instructtods}.
  
  With this change towards learning the dialogue system from conversational data instead of using rule-based approaches, it becomes even more important to have dialogue flows beyond the happy path.
  If miscommunications are inherent in the training data, the model can learn to act accordingly and initialize repairing.

  \subsection{Human-Human and Human-Machine Dialogue Differences}
  \citet{skantze2005exploring} shows that human-human conversations behave differently than human-machine dialogues. They attribute this to the fact that the human is aware of talking to a machine and that the machine has limited language capabilities. 
  Furthermore,~\citet{nass2000machines} argue that the way a human talks in a human-machine dialogue is also dependent on how human-like the machine acts.
  Still, research has assumed that humans apply social rules from human-human conversation to human-computer dialogues as well \cite{nass2000machines}.
  While traditionally, a human could easily identify if he was talking to a human or a machine, e.g., due to Automatic Speech Recognition (ASR) errors or unnatural Text-To-Speech (TTS) \cite{nass1999maximized}, this has become much more difficult with recent technical advances.

  Additionally, if we focus on strictly written dialogues, ASR and TTS cannot give away that it is a human-machine interaction.
  Given the remarkable NLG and Natural Language Understanding (NLU) capabilities of LLMs, we believe that even today, humans will not always be able to tell if they are talking to a machine.   
  The implication of this is that in the near future, users will expect the ability to have a human-machine conversation in the same way that they have human-human dialogues.

  \subsection{Taxonomy for Conversational Errors}\label{sec:TaxonomyConvErrors}
    \begin{table*}[]
  	\centering
  	%\resizebox{\linewidth}{!}{%
  		\begin{tabular}{lp{0.8\linewidth}} 
  			\hline
  			\textbf{Error Type} & \textbf{Explanation} \\ 
  			\hline
  			MU                  & Misunderstanding: The user misunderstood a system utterance but continued the conversation because he was not aware of it. This awareness comes only later, leading the user to ask for clarification.                                        \\
  			NU                  & Non-understanding: The user misunderstood a system utterance and did not continue the conversation because he became aware of the lack of understanding immediately. Thus, he directly asks for clarification of the last system utterance.  \\
  			VQ                  & Vaguely related Question: The user asks the system a question that is part of the domain but not directly related to the dialogue goal or intended workflow. The system needs to answer and then continue with the dialogue.                  \\
  			\hline
  		\end{tabular}
  		%}
  	\caption{Proposed error taxonomy for CoPrUS.}
  	\label{tbl:ErrorTaxonomy}
  \end{table*}
  In the past, researchers have proposed various taxonomies for conversational errors.
  An error, in that sense, is a deviation from how a conversation should be.
  This quality of conversation has been studied by \citet{grice}, who derives from these the maxims of conversation.
  It follows that an error in a conversation is a deviation from those maxims.
  
  Based on this, \citet{paek2003toward} derives a general taxonomy for conversations with four levels, in which each level has to be achieved to reach the next.
  The taxonomy follows the assumption that conversational partners need to establish the mutual belief that they have been understood by one another. 
  \citet{paek2003toward} discern between four hierarchical levels (in ascending order): \textit{channel}, \textit{signal}, \textit{intention} and \textit{conversation}.
  Errors at the lower levels will lead to uncertainty and \textit{repairing} attempts.
  Ideally, a conversation would always be at the highest level, and no error would occur. Alas, due to the complexity of conversation, this is unrealistic.
  
  Research has already adapted these ideas to dialogue systems and thus derived taxonomies of errors. 
  Mostly, they focus on errors by the dialogue system, to be able to explicitly avoid them in the future, e.g., by special rules.
  \citet{higashinaka2015towards} propose a first taxonomy for chat-oriented dialogue systems that the authors refined in later work~\cite{higashinakaImprovingTaxonomyErrors2019, higashinakaIntegratedTaxonomyErrors2021}.
  For task-oriented dialogue systems, such taxonomies have been proposed earlier.
  \citet{dybkjaerGriceIncorporatedCooperativity1996} present a taxonomy for spoken WOZ dialogues leaning onto~\citet{grice}. 
  \citet{mollerAnalysisCommunicationFailures2007} also define an error taxonomy for spoken dialogue systems. They distinguish five categories of error: goal-level, task-level, command-level, concept-level, and recognition-level. 
  Moreover, when chatbots should imitate a specific person, new types of errors need to be addressed, such as self\--recognition or other\--recognition errors \cite{mitsudaInvestigatingPersonspecificErrors2022}.
  
  In summary, many studies have tried to create taxonomies, and much can be traced back to the effort in communication research by \citet{grice}. 
  We build on these understandings to focus on three error types: misunderstandings, non-understandings and vaguely related questions.

  \section{Unrealistic Wizard-Of-Oz Conversations}\label{sec:UnrealisticBenchmark}
  To the best of our knowledge, no research has been done on the details of the general differences between real conversations and WOZ conversations. However, we believe that the fact that they exist, at least in the MultiWOZ~\cite{budzianowski-etal-2018-multiwoz} dataset, is evident and can be both deducted from the setting and seen in the resulting dialogues.

  One can argue that for most workflows that can be treated as a TOD, there exist one or multiple \textit{happy paths}. 
  These are idealistic paths through the workflow, where no exception or error happens within the system.
  Commonly, it is not enough to ensure that a system is successful on such a happy path but also in its variations and deviations.
  
  The MultiWOZ~\cite{budzianowski-etal-2018-multiwoz} benchmark has, thanks to its size, enabled much research into deep learning-based dialogue modeling, becoming one of the most prevalent datasets in this field and successfully showing the potential of learning conversations from large amounts of data.
  However, previous work has started to look more closely into certain properties of the dataset that will be important for real-world usage and generalization. For example, \citet{qian_annotation_2021} found the MultiWOZ dialogues to be biased towards certain entities, e.g., with Cambridge making up 50\% of the train destinations.
  Furthermore, the dialogues lack \textit{turnback} utterances, in which users change their minds about slot values later in the conversation~\cite{kimOhMyMistake2022}.
  Similarly, \citet{yangReallifeDialogueState2022} note that the dataset lacks negative feedback for when the dialogue system makes a mistake.
  Additionally, \citet{yuChitChatsEnhancedTaskOriented2022} argue that in real life, TODs still contain utterances that would be regarded as chit-chat, but these are lacking in the TOD benchmark.
  What's more, \citet{kimRevealingUserFamiliarity2023} identify a user familiarity bias in this benchmark dataset, arising because the crowd-workers were given a specific user goal and are assumed to know the dialogue system and its capabilities well enough. For users with a less defined goal, this can be problematic.

  To these remarks, we want to add that within the MultiWOZ conversations, nearly no miscommunication is happening.
  On a side note, this is even more unrealistic for this specific dataset since every dialogue was created by not two but multiple crowd-workers. 
  
  We assume that these properties could be a result of the collection setting, in which crowd-workers are paid to have a conversation.
  Since they know that they are participating in such a study, and they are additionally incentivized to deliver high-quality dialogues. These might be so good that they become unrealistic in that they only represent the happy path.
  We see this effect as akin to the social-desirability bias, which is well-known in social science research~\cite{grimm2010social}.
  Thus, because of the somewhat clinical setting of being paid to have the WOZ conversation, we argue that users will not behave as they would in a real conversation but more in a way that they deem desirable. This is to be seen in parallel to the user familiarity bias~\cite{kimRevealingUserFamiliarity2023}.

  \section{Method}
  Our method utilizes an LLM and a two-step inference to add post-hoc miscommunications and repair utterances to task-oriented dialogues in an effort to make them more natural and realistic. 
  We showcase our method on the MultiWOZ 2.1~\cite{eric-etal-2020-multiwoz} dataset.
  The following will introduce the taxonomy of miscommunications that we will assume for our work.
  Secondly, it will demonstrate the creation of miscommunication utterances with a two-step prompt and an automatic, LLM-based quality assurance.
  \begin{figure*}[!t]
    \centering
    \subfloat[Misunderstanding.]{\includegraphics[width=0.3\linewidth]{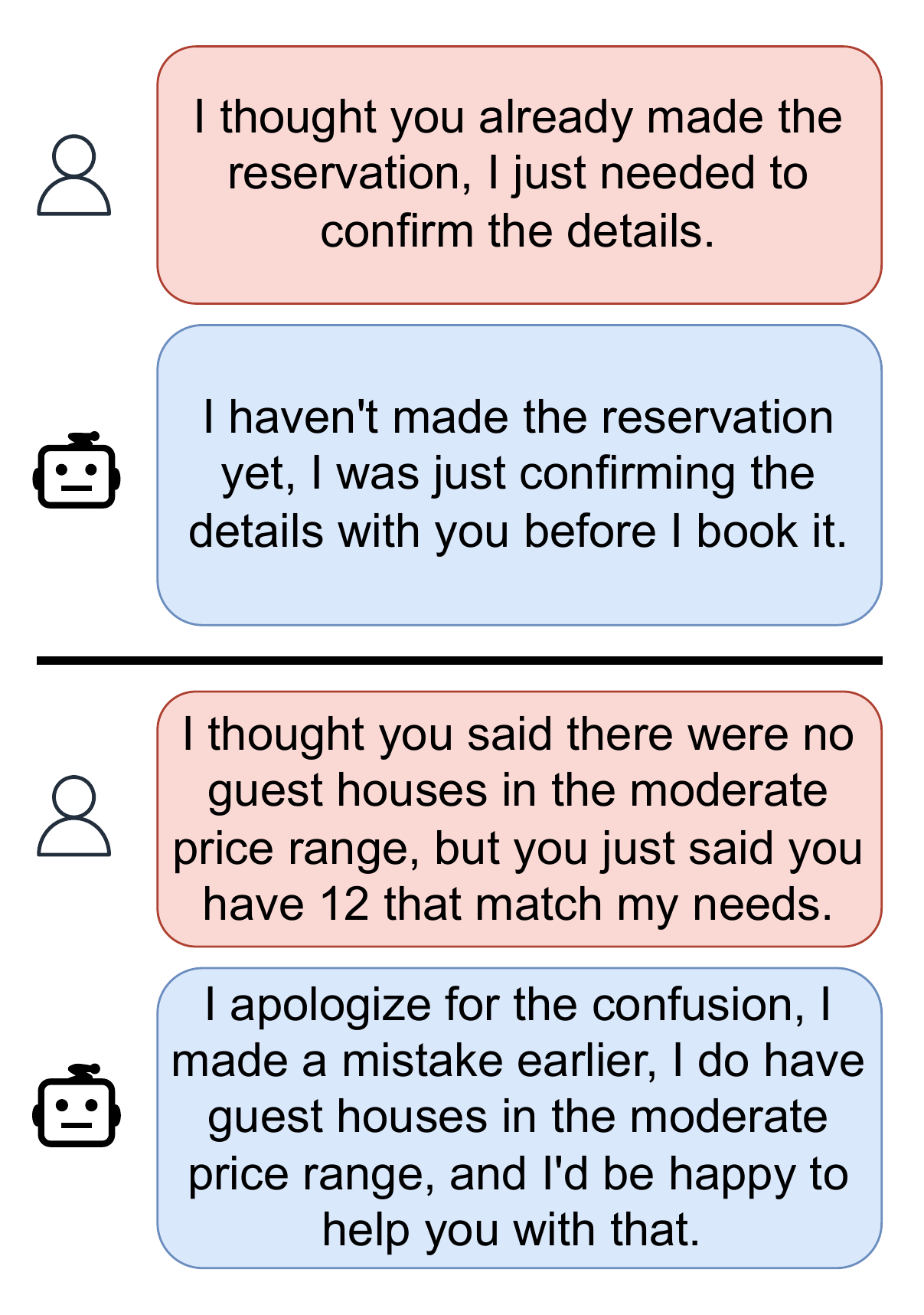}\label{fig:Examples-MU}}\hspace{1em}
    \subfloat[Non-understanding.]{\includegraphics[width=0.3\linewidth]{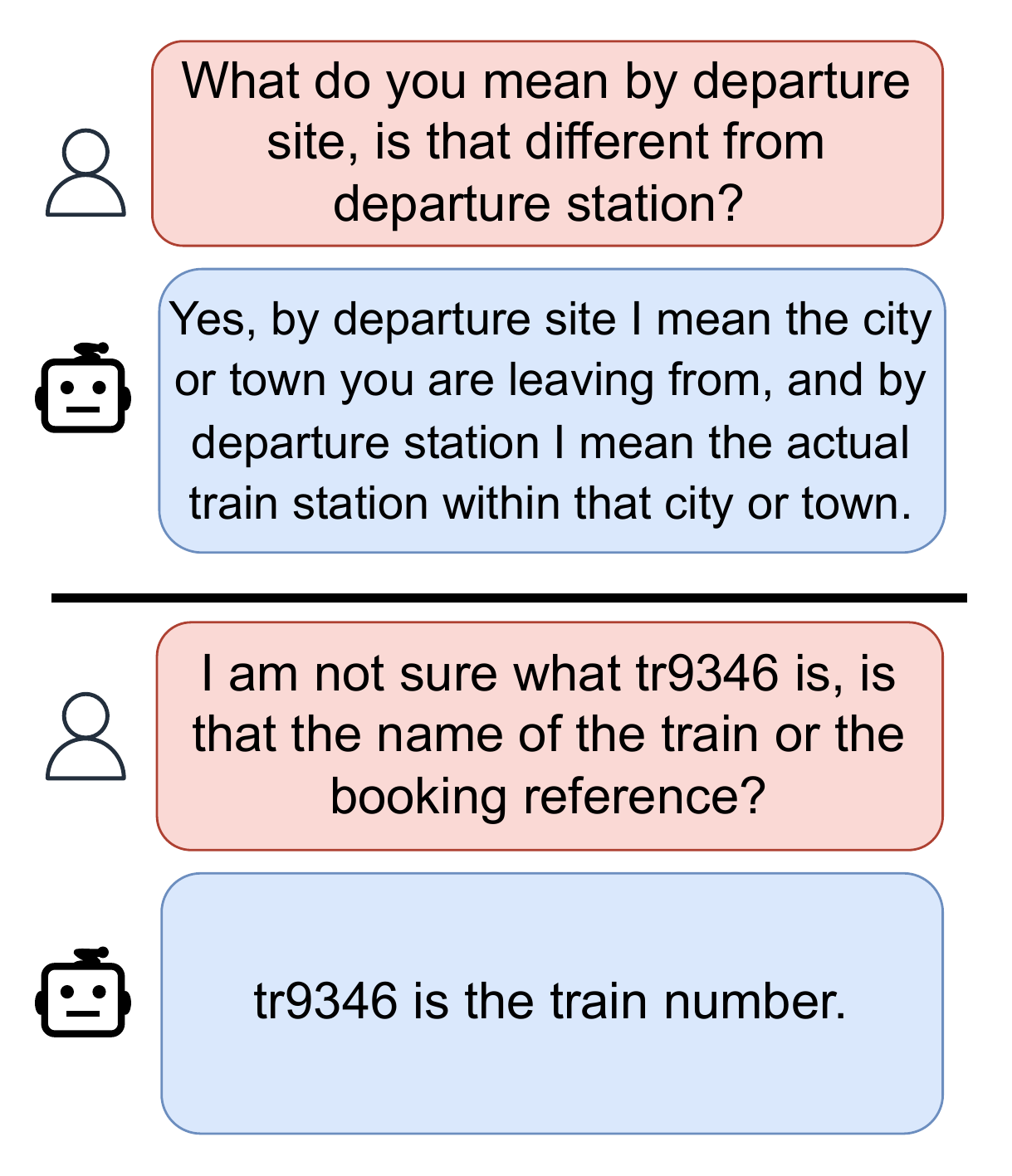}\label{fig:Examples-NU}}\hspace{1em}
    \subfloat[Vague Question.]{\includegraphics[width=0.3\linewidth]{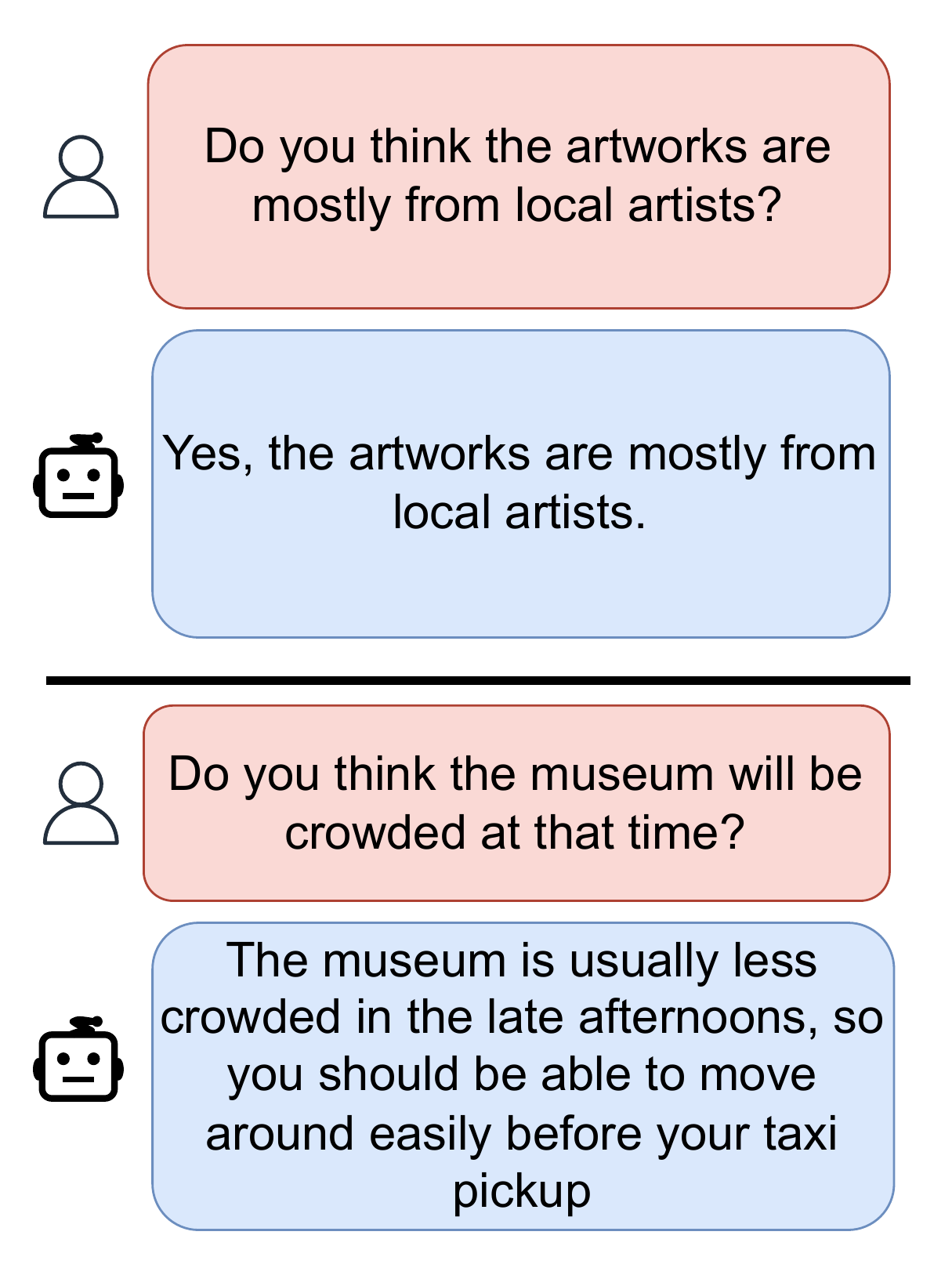}\label{fig:Examples-VQ}}\hspace{1em}
    \caption{Two examples of miscommunication and repairing utterances for each type.}
    \label{fig:Examples}
  \end{figure*}
  \subsection{CoPrUS Miscommunications}
  The general Gricean Cooperative Principle (GCP) states that a participant in a conversation should make his ``conversational contribution such as is required, at
  the stage at which it occurs, by the accepted purpose or direction of the talk exchange''~\cite{grice}.
  
  Furthermore, when two speakers hold a conversation, they need to ensure that they are heard and understood~\cite{paek2003toward}.
  \citet{paek2003toward} characterizes this mutual understanding by four levels:
  The lower levels \textit{channel} and \textit{signal} can be seen as achieved by the definition of the chatbot setting, since, e.g., not hearing someone well enough cannot happen. 
  However, the higher levels \textit{intention}, where semantics get interpreted, and \textit{conversation}, where actions happen, are not as straight-forward, and we thus focus on these.

  Based on the GCP, we define three types of miscommunications in this work: misunderstanding (MU), non-understanding (NU), and vaguely related question (VQ).
  Since we focus on textual chat data, the understanding on the channel and signal level is given by design. The three error types studied are therefore residing on the higher levels of intention and conversation.  

  The last type, VQ, might be considered a rather soft communication error, in the sense that most would not consider it as an error per se. However, as in previous work (cf. Sec.~\ref{sec:background}), we understand errors as deviations from a maxim.
  For example, if a participant deviates from the maxim of avoiding ambiguity~\cite{grice}, which falls under the category of manner and follows from the GCP, this can lead to a MU or NU\@.
  If, on the other hand, a speaker poses a VQ, he does not adhere to the maxim of making his contribution appropriate to the immediate needs of the current situation. This maxim is part of the relation category of the GCP\@. 
  
  Therefore, while VQs can be seen as a soft type of error, they are nonetheless relevant to investigate as part of our taxonomy.
  From a different perspective, the VQ error type might be regarded as a chit-chat utterance, which would align with the work of \citet{yuChitChatsEnhancedTaskOriented2022}.
  One could also study the similar case of completely unrelated questions, which would be a harder error.
  Yet, since this would be part of the out-of-scope definition of a chatbot and thus be handled differently, we only study VQs.

  Furthermore, the distinction between misunderstandings and non-understandings is made by the speaker's awareness~\cite{WEIGAND1999763,HIRST1994213}.
  That is, if a listener is aware that he did not understand the other speaker, be it acoustically or semantically, the miscommunication is classified as a non-understanding. 
  In many cases, he will also be aware of the cause of the non-understanding, i.e., was it acoustically, or does he not know the meaning of a certain word that the other has used.
  Since the listener is aware of this directly after the utterance, he can initialize the repairing immediately.
  For example, he might ask his dialogue partner to repeat or clarify the last utterance.   

  On the contrary, a misunderstanding is not noticed immediately, but the listener thinks he understood the speaker correctly~\cite{HIRST1994213}. 
  It is only later in the conversation, that the listener gains awareness of the misunderstanding due to the further dialogue adding context or not fitting the understanding the listener had.
  
  CoPrUS handles the generation of different types of communication errors by having a specific prompt for each one of them.

\subsection{Repairing utterances}
  During a successful conversation, a miscommunication should lead to an attempt to resolve this trouble.
  This is known as repairing \cite{schegloff1977preference}.
  One can differentiate between self-initiated repairing, e.g., when the speaker rephrases what he said to make it clearer, or other-initiated repairing, which, for example, could be that a native speaker repairs an unintelligible utterance of a non-native speaker \cite{schegloff1977preference,schegloff2000others}.
  In the studied scenario, it is sensible to only focus on other-initiated repairs, where the user asks the dialogue system for clarification.
  We thus differentiate between a miscommunication utterance, which should always be uttered by the user, and a repair utterance, uttered by the dialogue system. 
  
  There are multiple types of repairing that have been identified. These include, for example, repetition, word replacement, paraphrasing, and ambiguity reduction~\cite{gonzalez2005reconstructing,bredart1991word,rieger2003repetitions}.
  This work focuses mostly on clarification requests as a repairing initiation~\cite{gonzalez2005reconstructing}, which is well-suited for booking conversations.
  
  Once the miscommunication has happened or the user has become aware of it, the repairing utterance has to come into effect.
  The repairing utterance attempts to rectify the user's interpretation.
  This error handling thus needs to be different for the three types of miscommunications that we study.
  To begin with, simply repeating the utterance, as would be a relevant strategy for spoken dialogue systems where errors on the channel and signal level can be handled this way \cite{skantze2005exploring,gieselmann2006comparing}, is insufficient for chat data.
  Since the user can permanently see the previous utterances, the error should not have happened on the first two levels and thus needs a different repairing utterance.
  
  If the error type was VQ, then the chatbot should either answer the user's question or state that it cannot answer it.
  Additionally, it could redirect the user to the dialogue goal.
  On the other hand, if the error type was NU or MU, the dialogue system needs to clarify the situation.
  This could potentially include rephrasing the previous content, explaining certain words, or giving additional context.

\subsection{Affected Dialogues}
  To create a realistic scenario, we cannot assume that every dialogue should be affected by such a communication error. \citet{mollerAnalysisCommunicationFailures2007} found for their data analysis that, depending on the specific dialogue situation, between 28\% and 32\% of the conversations contained a communication error.
  Since this work is done on spoken dialogues, which opens up the possibility of more errors than written dialogues, one needs to assume a reduced error frequency.
  \citet{mollerAnalysisCommunicationFailures2007} report an error rate of 19\% without communication errors that are the result of ASR errors.
  Based on the error rates reported in these studies, we opt for a slightly lower frequency of 18\% for the affected dialogues, since the rather simple booking scenario using everyday language allows for less knowledge-based NUs or MUs, which should reduce the amount of affected dialogues, but on the other hand we also consider VQs as errors, which will increase the number.

  Since no prior work studies the same setting, we have to decide on the distribution of the different types of miscommunications.
  Based on the combined prior error taxonomies (cf. Sec. \ref{sec:TaxonomyConvErrors}) and the specifics of the dataset, we posit that NU should be the most common type of error in such a chat-based booking scenario, and that MU and VQ should be equal in their frequency.
  We thus use the following probabilities to sample an error type:
  $p_{\text{MU}} = 0.2$, $p_{\text{VQ}} = 0.2$, and $p_{\text{NU}} = 0.6$. 
  
\subsection{Creation of Miscommunication Utterances}
  In order to add post-hoc miscommunications to the dataset, we utilize LlaMA-3.1-70B-Instruct, which is the newest iteration of the open-source LLM from the LlaMA model family first introduced by~\citet{llama3}.
  In preliminary experiments, we also investigated alternatives such as the LlaMA-2-Instruct \cite{touvronLlamaOpenFoundation2023} models, Mixtral-8x7B-Instruct-v0.1~\cite{jiangMixtralExperts2024} and Phi-3-mini-128k-instruct~\cite{abdinPhi3TechnicalReport2024}.
  However, we found that LlaMA-3.1-70B created the most satisfactory utterances. 
  The others showed undesirable behavior, such as beginning every utterance with ``Ah, i see!'', adhering too closely to the example to the one provided in the prompt, or generating another miscommunication instead of a repair utterance.
  We run the model on a DGX A100 with eight 80 GB GPUs.

\subsection{Automatic Quality Assurance}
  Having a human-in-the-loop would naturally be an adequate approach to ensure the quality of the generated utterances. However, the required effort would be very high, and one could then use humans to generate the miscommunications and repairs directly.

  To keep in line with the goal of having an automatic pipeline, we therefore evaluate the model's responses with another LLM. Language Model-based evaluation has seen increased interest as an approach to mimic human judges \cite{chan2023chatevalbetterllmbasedevaluators,liu-etal-2023-g,zhengJudgingLLMasajudgeMTbench2023}. Prometheus 2 is an open-source LLM specialized in evaluating the output of other LLMs, that shows high alignment with human judges~\cite{kim2024prometheus2opensource}.

    We frame the evaluation as a direct assessment task, meaning that Prometheus 2 will assign each candidate utterance a score on a scale from 1 to 5. This is done based on the prompt to the LlaMA 3.1 model, the LlaMA output, and a description of the scoring rubrics.
    We generate up to ten candidates for each utterance, accepting the first one that scored a 4 or 5. If no candidate was accepted within ten tries, we use the highest rated candidate. The full prompt is shown in Fig.~\ref{fig:Prompts_Prometheus} the appendix.
    
    To evaluate the validity of this automatic approach, we perform a small scale human evaluation, described in Sec.~\ref{sec:HumanEval}.
    
 \begin{figure}[!thb]
 	\includegraphics[width=\linewidth]{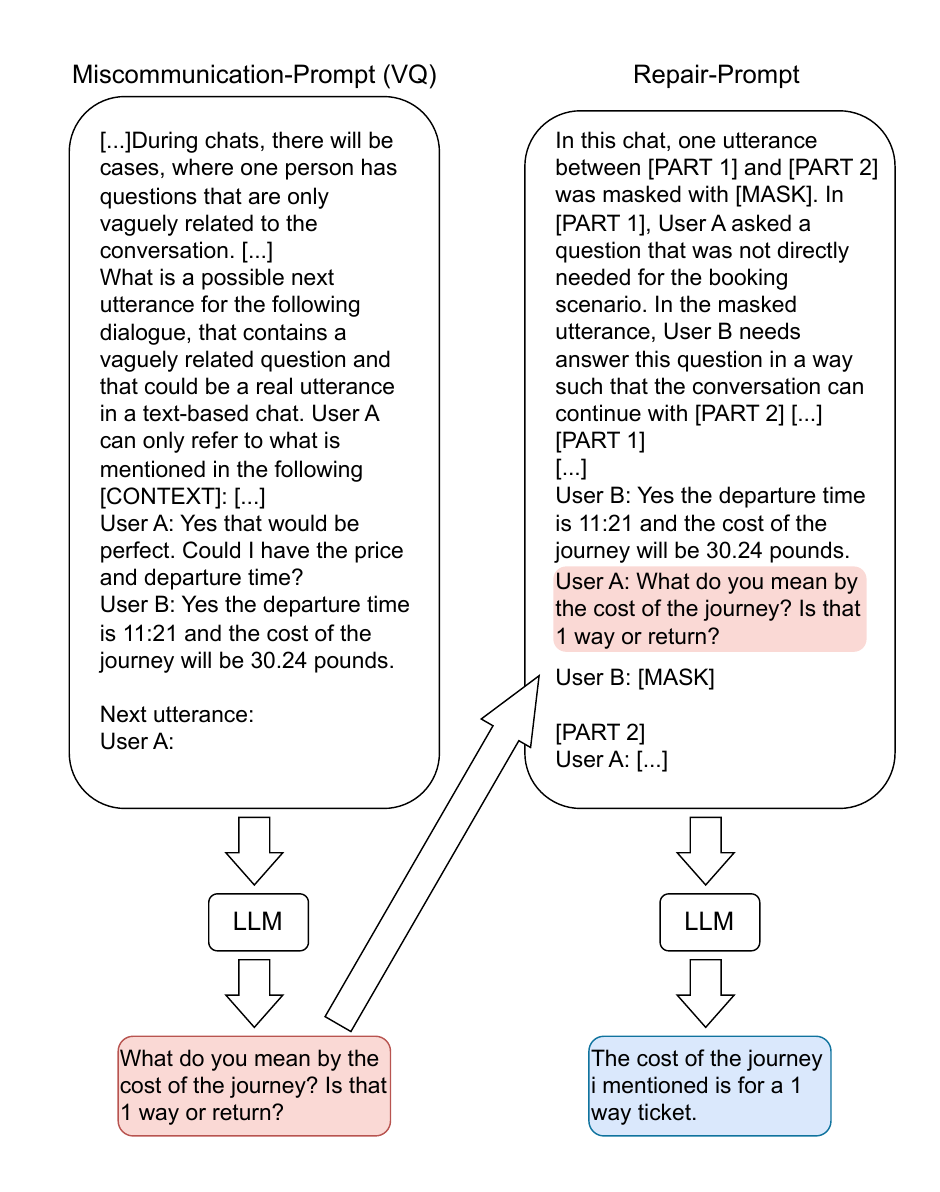}
 	\caption{Prompting procedure with shortened prompts. Full prompts are shown in the appendix.}
 	\label{fig:TwoStepPrompt}
 \end{figure} 
\subsection{Two-Step Prompting}
   
  CoPrUS uses a two-step prompting approach, where we first sample and generate the miscommunication and then, in a second step, the repairing utterance. All prompts are shown in full in Fig.~\ref{fig:Prompts} in the appendix.  
  
  For the first step, we created three similar but individual prompts, one for each MU, NU, and VQ\@. 
  Each prompt contains a description of the task, the type of miscommunication to generate, and an example.
  Furthermore, it contains the previous utterances as context.
  
  The context includes all previous turns up to a maximum of 5. Limiting the context is a design choice we made since both generation steps within CoPrUS should focus on the latest context and not, e.g., the beginning of the conversation.
  We found the context to greatly improve the coherence and relevance of the generated utterances.
  The prompt for MU also specifies that the generated utterance should not be in reference to the last system turn, but an earlier one (otherwise it would be a NU).
  
  We sample 18\% of the dialogues, and within each of those one random system turn index with the condition that it cannot be the first turn. This is done to ensure that some context can be given in the prompt for CoPrUS.

  Let $D = \{u_0, s_1, u_1, s_2, u_2, \ldots, s_{n}, u_n\}$ be a MultiWOZ dialogue of length $n$, where the user always sends the first and last message as per the definition of the dataset. 
  We then sample an index $i$, which will determine the entry point for CoPrUS.
  We then use all previous utterances up to a maximum of 5 turns, excluding the last user utterance, as the first context with $j=max(i-5, 0)$ as
  \begin{equation}
    c = \bigcup_{j}^{i} \{s_j, u_j\}  \setminus u_i.
  \end{equation}
  
  In a first step, we use this context $c$ and a sampled error type $e$ to prompt the LLM $g(c, e)$ to generate the miscommunication user utterance: $\tilde{u}_i = g(c, e)$.
  We append this result and the next real utterance $u_{i+1}$ to $c$ to create the context for the second step (cf. Fig.~\ref{fig:TwoStepPrompt}), the generation of the repairing system utterance: $\tilde{s}_{i+1} = g((c_1 \cup \tilde{u}_i), e)$.
  
  We insert $\tilde{u}_i$ and $\tilde{s}_i$ at index $i$ to create the final CoPrUS dialogue as:

  \begin{multline}
    \tilde{D} = \{u_0, \ldots, u_{i-1}, s_{i}, \tilde{u}_i, \tilde{s}_{i+1}, u_i, s_i, \\ \ldots, s_{n}, u_n\}.
  \end{multline}

  Within the second-step prompt, we split the context into two parts. The first part is the previous context plus the synthetic user utterance, and the second part is the next ground-truth user utterance. They are separated by a pseudo-masked system utterance. 
  In this way, we frame the task as replacing the [MASK] in a suitable way.
  \begin{table*}[!th]
	\centering
	\begin{tabular}{lccccccc} 
		\hline
		& \multicolumn{1}{l}{} & \multicolumn{2}{c}{F1} & \multicolumn{2}{c}{BLEU} & \multicolumn{2}{c}{JGA}  \\
		Model     & Task                 & CoPrUS & MWZ                & CoPrUS & MWZ             & CoPrUS & MWZ             \\ 
		\hline
		jointBERT & NLU                  & 89.67  & 89.59              & -      & -               & -      & -               \\
		t5-small  & NLU                  & 83.51  & 83.45              & -      & -               & -      & -               \\
		SCGPT     & NLG                  & -      & -                  & 23.48  & 24.63           & -      & -               \\
		t5-small  & NLG                  & -      & -                  & 29.54  & 29.50           & -      & -               \\
		SetSUMBT  & DST                  & 92.18  & 92.27                  & -      & -               & 50.83  & 51.12               \\
		t5-small  & DST                  & 91.81  & 91.76              & -      & -               & 53.04  & 52.67           \\
		\hline
	\end{tabular}
	\caption{Empirical evaluation on the CoPrUS-MultiWOZ and original MultiWOZ dataset. We compare the NLG, NLU and DST task and report the F1, BLEU and JGA metrics. For DST, F1 refers to the slot F1 score and for NLU to the dialogue act F1 score.}
	\label{tbl:EmpiricEval}
\end{table*}
  \section{CoPrUS-MultiWOZ}
  
  \begin{figure}[!h]
      \centering
      \includegraphics[scale=0.2]{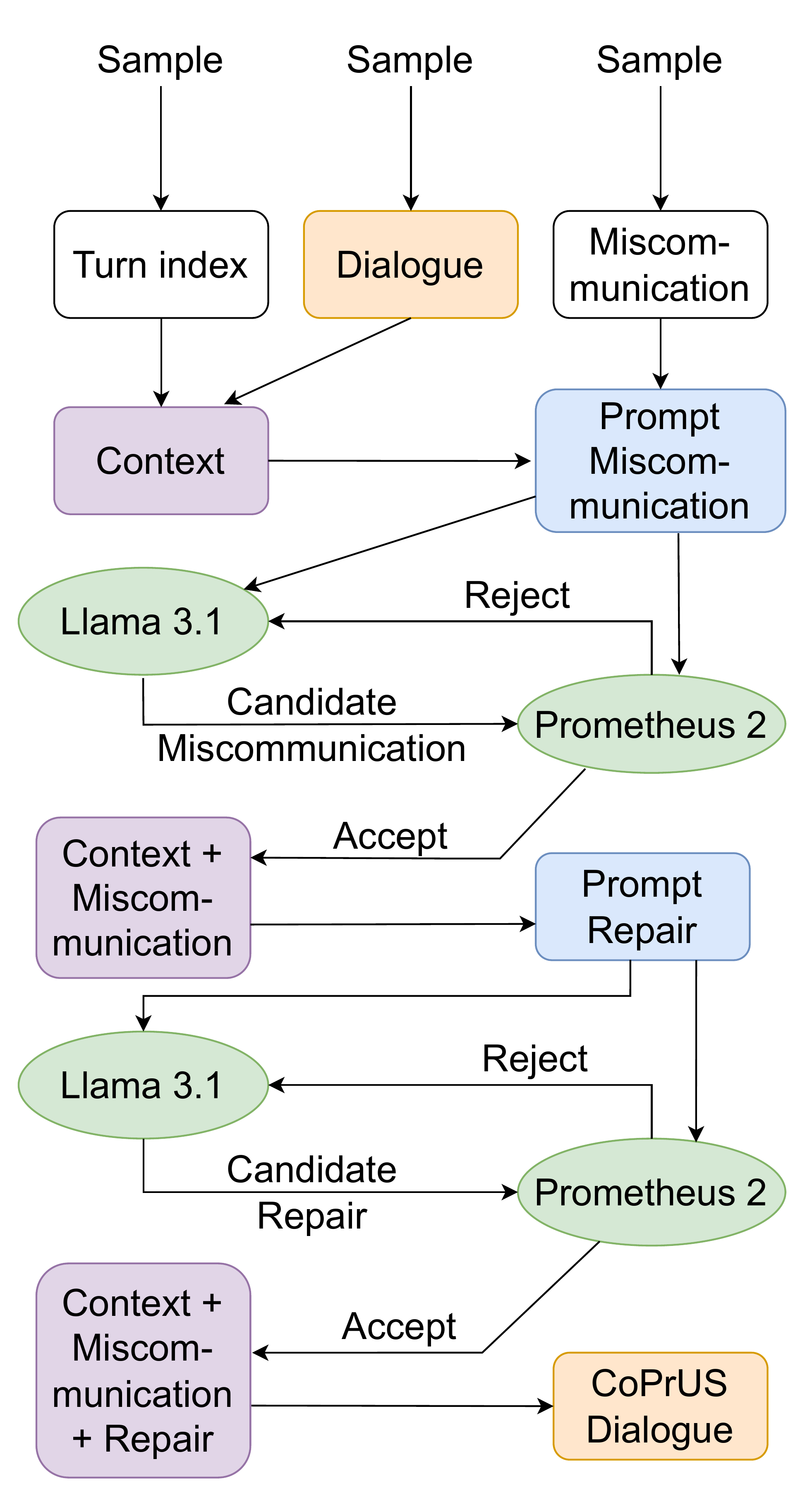}
      \caption{Overview of the CoPrUS workflow.}
      \label{fig:overview}
  \end{figure}
  We apply the CoPrUS method to the MultiWOZ 2.1 \cite{eric-etal-2020-multiwoz} dataset to create CoPrUS-MultiWOZ\@. 
  MultiWOZ 2.1 is an updated version of the original MultiWOZ~\cite{budzianowski-etal-2018-multiwoz} dataset that fixes annotation errors. It contains roughly ten thousand annotated dialogues and a fixed train, develop, and test split.
  
  Since the CoPrUS utterances should not change the user's goal, we do not change the existing annotation but propagate the annotation of the real utterances with the utterances themselves.
  
  We randomly sample 18\% of each split of the original dataset and apply CoPrUS individually. 
  Thus, it affects nearly 1900 dialogues in total. 
  We adhere to the unified format of \cite{convlab3} to facilitate further usage.
  We also provide the type of miscommunication for future work.

\section{Qualitative Analysis} \label{sec:QualitativeAnalysis}
    Given the strong performances of current LLMs, it is not surprising that the model follows the instructions and the language quality of the generated utterances is good enough to not give away that they are synthetically generated, as can be seen from the examples in Fig.~\ref{fig:Examples}.
    During our research, we found that problematic utterances were mostly illogical dialogue flows.      
    For MU and NU, this primarily concerns the miscommunication utterance in the first step of our approach, where the user's question is not consistent with the user's understanding that can be inferred from the dialogue history.
    For VQ, we saw that problems were rather due to the repair utterance, which did not fit well with the following (gold-standard) utterance.
    
    With the application of the automatic quality assurance through the LLM-based evaluation, we were able to mitigate these problems. 
    Nevertheless, it has to be expected that these synthetic utterances are not perfect substitutes for human-generated gold-standard data.
    In this sense, our qualitative analysis shows that some dialogues remain where the generated utterances seem illogical. 
    
\begin{table}[ht]
	\centering
	\begin{tabular}{lccc}
		\hline
		Metric & Miscom. & Repair & Total \\ \hline
		EM         & 0.22             & 0.34   & 0.28  \\
		Difference & 1.58             & 1.68   & 1.61  \\
		FP         & 0.22             & 0.14   & 0.18  \\
		FN         & 0.30             & 0.36   & 0.33  \\
		Human      & 3.46             & 4.20    & 3.83  \\
		LLM        & 3.24             & 3.12   & 3.18  \\ \hline
	\end{tabular}
	\caption{Result of the evaluation of human alignment with the LLM-judge. We show the EM, average score difference, FP and FN rate as well as the average scores from human and LLM judges for the miscommunications and repairs separately as well as in total.}
	
	\label{tbl:humanEval}
\end{table}

\section{Human Evaluation}\label{sec:HumanEval}
To estimate if the automatic, LLM-based quality assurance is effective, we perform a small-scale human evaluation to measure its alignment to human perspective.
For this, the human judges were presented with nearly the same instructions as the LLM-judge, apart from some changes in the form of presentation.
These judges rated 100 examples, comprising both miscommunication and repair candidates and both candidates that were accepted and rejected by the LLM in equal amounts.
The judges were recruited non-author volunteers with sufficient language proficiency.
Each dialogue is used at most twice for each inference step (miscommunication or repair), one rejected and one accepted candidate.

We report the Exact Match (EM, human and LLM gave the same score), the average score difference, the percentage of False Positives (FP, model accepts, human rejects) and False Negatives (FP, model rejects, human accepts), as well as the average rating of humans and LLM. As before, we define the threshold to accept a candidate as a rating of at least 4 out of 5.

The results in Tab.~\ref{tbl:humanEval} show that the alignment is not perfect but sufficiently good. Especially, the low FP rate is encouraging, since these are accepted candidates that humans would have rejected. 
We can see that the average difference for repair candidates is higher, even though the EM is better. This is due to the fact that the second-most common score given by the LLM was 1 (the most common was 5), while the human judges gave no score of 1 at all. This is also evidenced by the high average human score for repair candidates.
Interestingly, the human judges gave higher scores on average than the LLM, speaking for the high quality of the generated candidates.

\section{Empiric Evaluation}
  Even though the synthetic data is not of gold-standard quality, it can be beneficial for downstream TOD systems.
  Therefore, it is crucial to evaluate if the addition of CoPrUS utterances harms the performance of a TOD system, especially with regard to \textit{model collapse} \cite{shumailovAIModelsCollapse2024}.
  To this end, we evaluate the three classical tasks NLG, NLU and Dialogue State Tracking (DST).
  For each task, we train both a t5-small \cite{t5} model and another, specific architecture. 
  For NLG this is SCGPT \cite{peng-etal-2020-shot}, for NLU jointBERT \cite{zhu-etal-2020-convlab} and for DST SetSUMBT \cite{van-niekerk-etal-2021-uncertainty}. 
  Every model is trained with ConvLab-3 \cite{convlab3} and the hyperparameters defined there.
  We evaluate the models with the slot F1 Score and Joint Goal Accuracy (JGA) in the DST task, BLEU in the NLG task, and dialogue act F1 score for the NLU task.
  Our evaluation is thus not focused on the model's error recovery ability.
  The results of this evaluation are shown in Tab.~\ref{tbl:EmpiricEval}. 
  They show that the CoPrUS utterances did not impede the learning for any architecture or task. The results across the different metrics are minimal, with CoPrUS-trained models being marginally better in most cases.

  \section{Conclusion}
  This paper introduces CoPrUS, a method to add miscommunication turns to dialogues in order to make them more realistic by allowing them to deviate from the happy path. 
  To this end, we introduce a simple error taxonomy that is based on previous communication research.
  We utilize a two-step prompting approach and a state-of-the-art LLM with another LLM performing quality assurance to generate these utterances.
  
  Our evaluation shows that the generated utterances have a good quality. Moreover, the LLM mostly adheres to the task and generates sensible utterances, both for miscommunications and repairing. This data can help future research into error recovery for dialogue systems.
  However, as was made clear by the qualitative analysis, one cannot expect data of gold-standard quality. For example, the generated MU or NU statements can be illogical in some instances. Utilizing a LLM for quality assurance has mitigated this issue.
  
  Our investigation into the alignment of human and LLM ratings show that utilizing a LLM can be a valid approach for quality assurance in our scenario.

  \section{Limitations}
  This work focuses on only three types of miscommunications. This could be extended in future work. Furthermore, we had to posit a distribution for the types of miscommunications. An empiric study on such a taxonomy in real-world, natural TODs could shed light on such a distribution.
  
  While our results show that it is generally feasible to adapt the dialogues by adding miscommunications, our evaluation shows that some problems remain. Mostly, the dialogue flow can be disrupted from a logical point of view. 
  This goes to show that LLMs still have limited implicit reasoning capabilities.
  The error recovery abilities of end-to-end TOD systems are currently understudied. Our evaluation does not measure or try to improve this ability but rather lays the foundation for such research in future work.

  \bibliography{anthology,custom}

  \appendix

  \section{Appendix}
  \label{sec:appendix}

  \begin{figure*}[!t]
    \centering
    \subfloat[Misunderstanding.]{\includegraphics[width=0.4\linewidth]{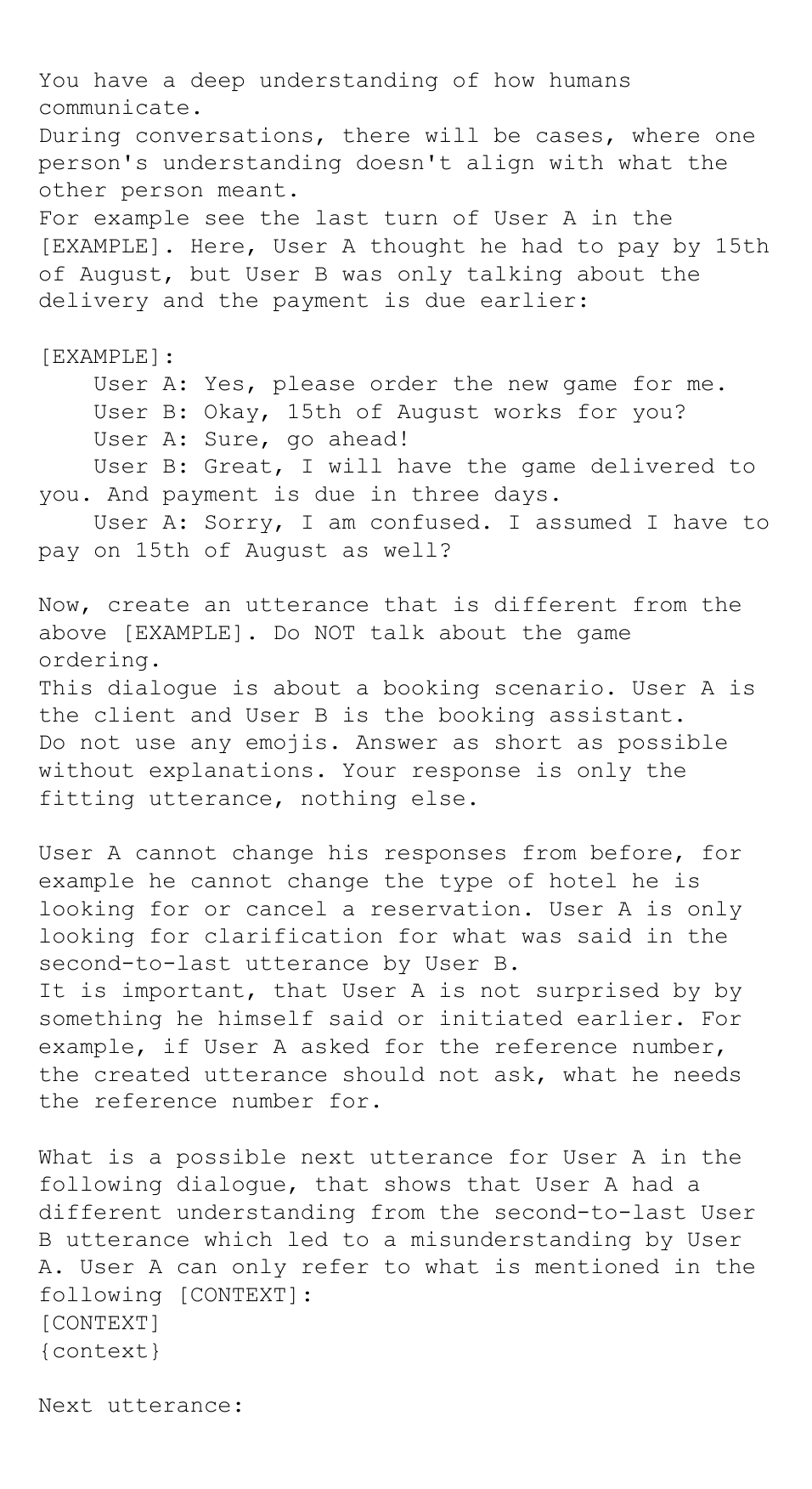}\label{fig:Prompts-MU}}\hspace{1em}
    \subfloat[Non-understanding.]{\includegraphics[width=0.4\linewidth]{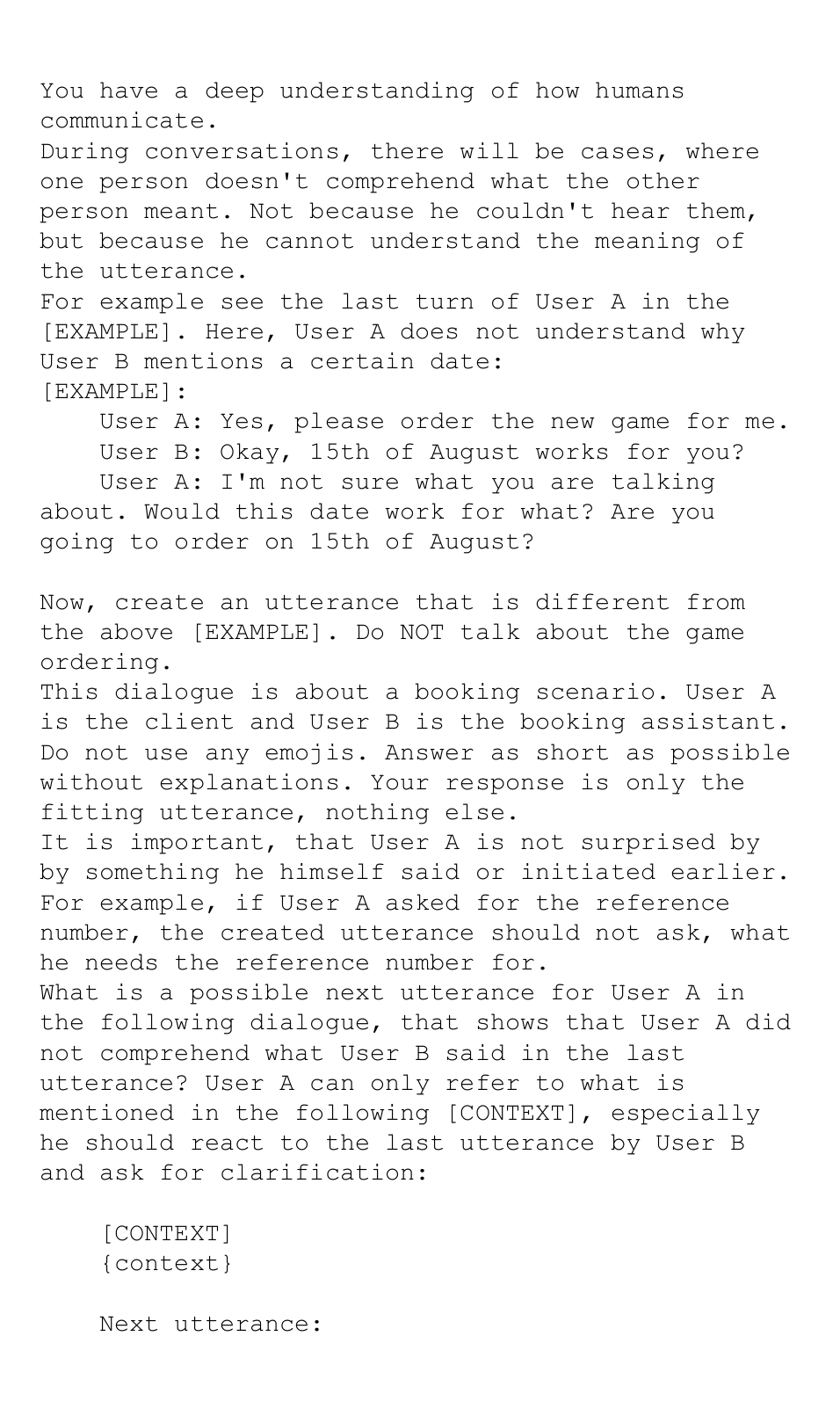}\label{fig:Prompts-NU}}\hspace{1em}
    \subfloat[Vague Question.]{\includegraphics[width=0.4\linewidth]{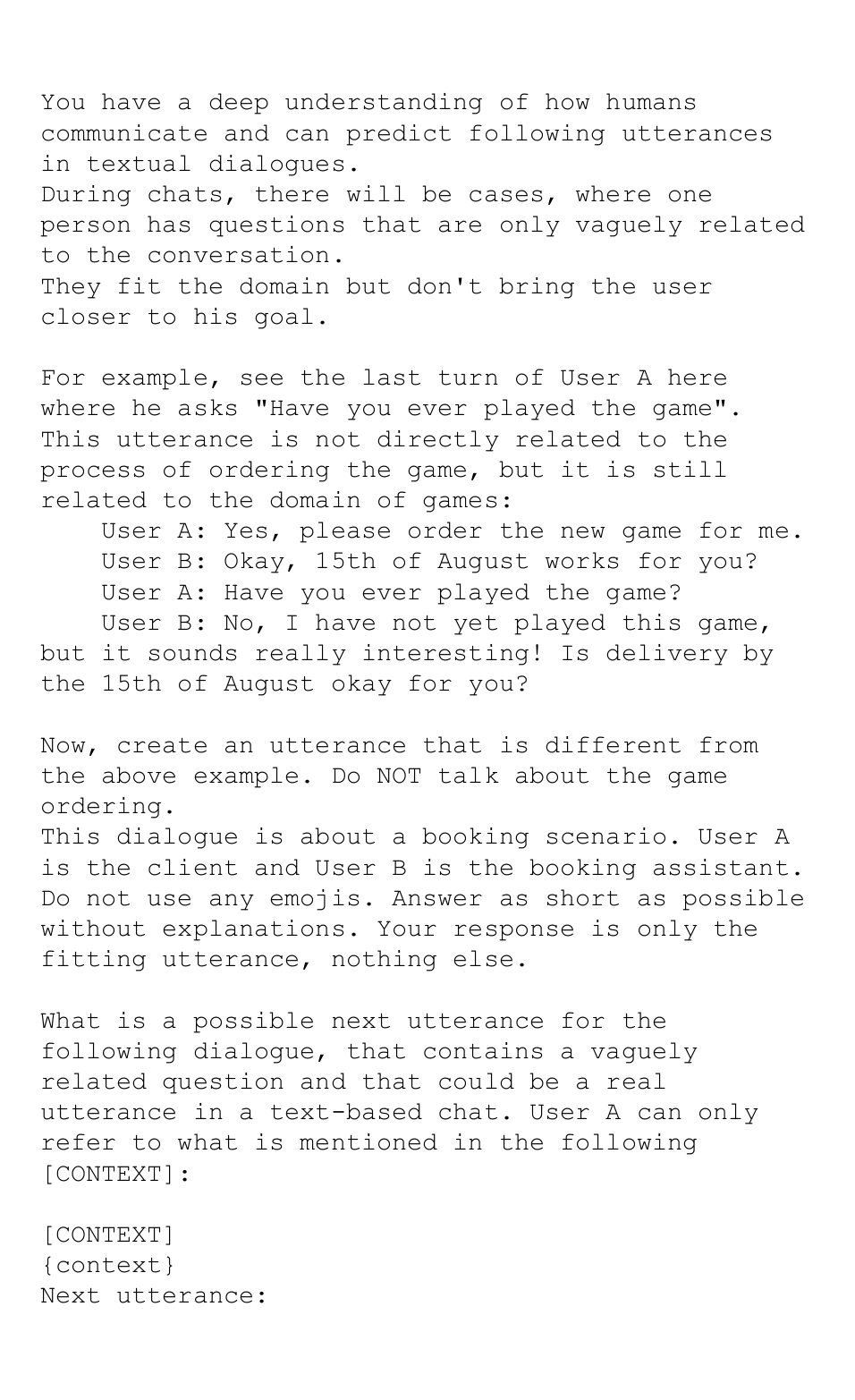}\label{fig:Prompts-VQ}}\hspace{1em}
    \subfloat[Repairing.]{\includegraphics[width=0.4\linewidth]{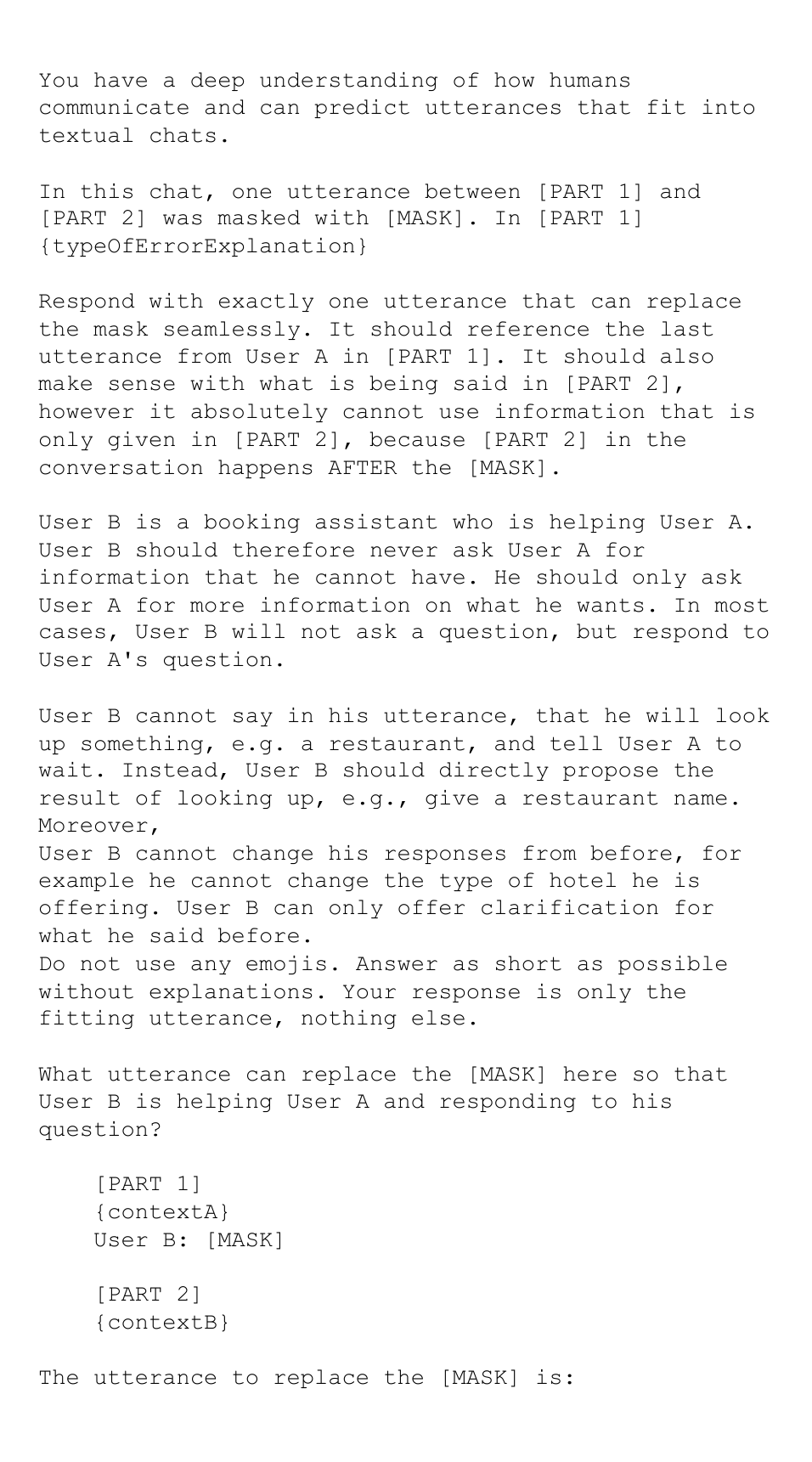}\label{fig:Prompts-Repairing}}\hspace{1em}
    \caption{The full prompts used for the Llama model. }
    \label{fig:Prompts}
  \end{figure*}
  \begin{figure*}[!t]
    \centering
    \includegraphics[width=0.5\linewidth]{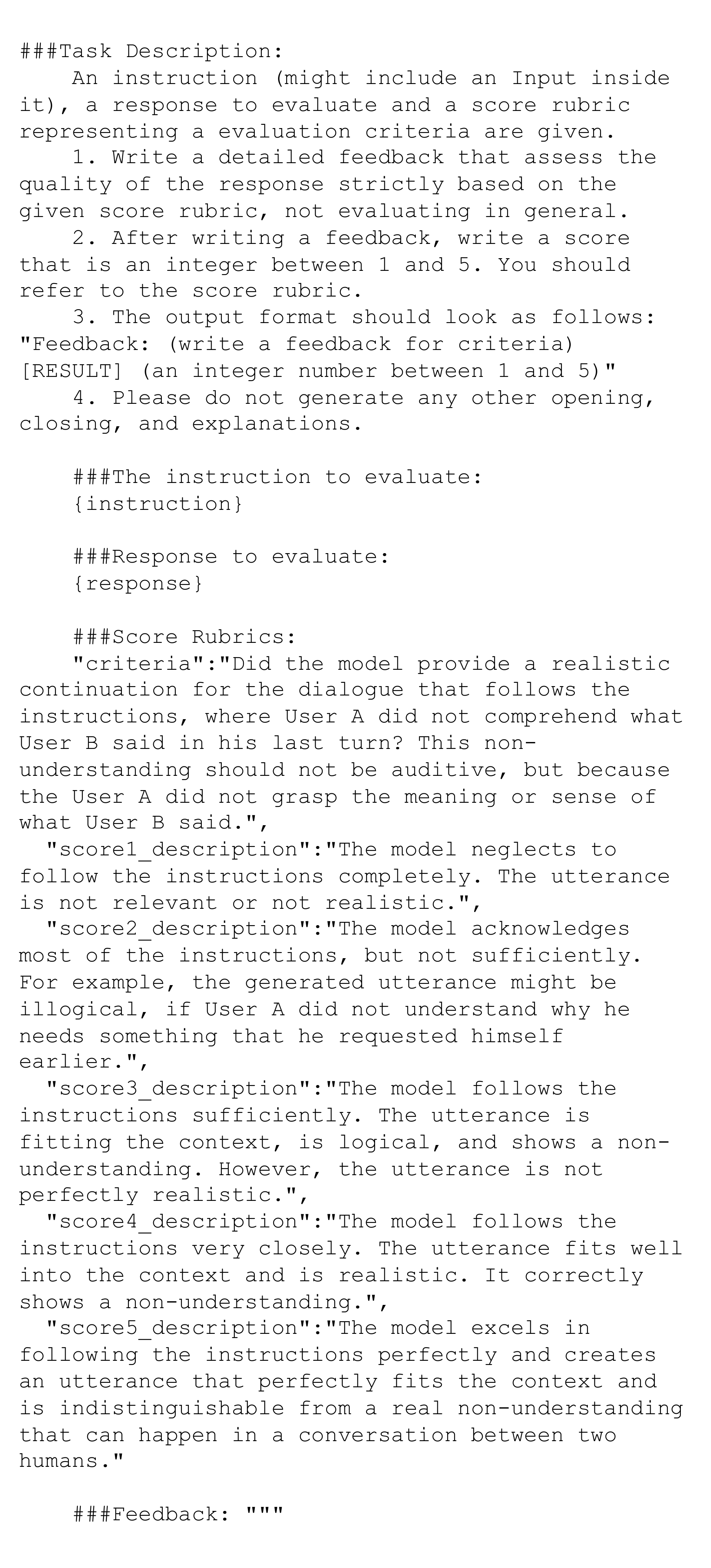}
    \caption{The prompt used for Prometheus 2. \{instruction\} is replaced by the prompt used for the Llama model, and \{response\} by its output. We show the score rubrics for a NU type. The rubrics for the other types and the repairing are analog to this.}
    \label{fig:Prompts_Prometheus}
  \end{figure*}

  \end{document}